# Undersampling and Bagging of Decision Trees in the Analysis of Cardiorespiratory Behavior for the Prediction of Extubation Readiness in Extremely Preterm Infants*

Lara J. Kanbar[+], *Student Member, IEEE*, Charles C. Onu[+], *Student Member, IEEE*, Wissam Shalish, Karen A. Brown, Guilherme M. Sant'Anna, Doina Precup, and Robert E. Kearney, *Fellow, IEEE*

*Abstract*— Extremely preterm infants often require endotracheal intubation and mechanical ventilation during the first days of life. Due to the detrimental effects of prolonged invasive mechanical ventilation (IMV), clinicians aim to extubate infants as soon as they deem them ready. Unfortunately, existing strategies for prediction of extubation readiness vary across clinicians and institutions, and lead to high reintubation rates. We present an approach using Random Forest classifiers for the analysis of cardiorespiratory variability to predict extubation readiness. We address the issue of data imbalance by employing random undersampling of examples from the majority class before training each Decision Tree in a bag. By incorporating clinical domain knowledge, we further demonstrate that our classifier could have identified 71% of infants who failed extubation, while maintaining a success detection rate of 78%.

## I. INTRODUCTION

Respiratory management of extremely preterm infants (birth weight ≤ 1250g) is challenging. These infants are born with underdeveloped lungs and immature control of breathing. As a result, they often require endotracheal intubation and invasive mechanical ventilation (IMV) during the first few days of life [1].

IMV is a life-saving therapy, but when used for a long period, protracted IMV is associated with increased morbidities and mortalities [2]. As such, timely and effective weaning and extubation is critical since reintubation is technically difficult and has been associated with adverse effects such as lung trauma, infection, lung collapse, and death [3, 4].

To assess extubation readiness, clinicians routinely use blood gas results, ventilator settings, and clinical expertise. However, there is no consensus on an objective weaning protocol, and practice varies individually and amongst institutions [5]. Indeed, studies have reported wide ranges of reintubation rates depending on several factors including the time frame used to define extubation failure [4, 6]. In this paper, extubation failure is defined as the need for reintubation within 72hrs of extubation.

Objective predictors of extubation readiness have been investigated using clinical variables, breath measurements of respiratory components, minute ventilation tests, and spontaneous breathing trials [7-10]. Most of these evaluations are time-consuming and require manual analysis of respiratory signals, limiting the analysis to small subsets of infants within a single institution. Therefore, the Automated Prediction of Extubation Readiness (APEX) study was designed as a multi-center, interdisciplinary study to predict extubation readiness in these infants using objective, automated measures of cardiorespiratory behavior. Our previous work on APEX involved a pilot study of 53 patients recruited at one institution [11].

This paper explored the predictive capability of quantitative measures of cardiorespiratory behavior to determine extubation readiness in a multi-institutional population. In our previous work, heart rate and respiratory variability provided valuable insight into the extubation outcome [8]. Within this larger multicenter infant population, we used scalar cardiorespiratory variability measurements and accommodated two key classification challenges. First, our dataset contained a relatively small number of examples compared to the number of features being evaluated. Second, the dataset suffered from severe class imbalance: the ratio of success to failure cases was 85:15. Therefore, it was crucial to use careful feature selection to avoid overfitting and adopt appropriate methodology to account for class imbalance during training.

We developed classifiers based on "bags" of decision trees, called Random Forests. Random forests address the first challenge of feature selection in an embedded fashion: the process of building individual trees identifies features that split the examples well. To address class imbalance, we applied random undersampling of the majority class before training of each decision tree in the bag of classifiers. This classifier structure was first presented in the literature as Balanced Random Forests (BRF) [12]. We demonstrate empirically that the use of BRF leads to a classifier with a more balanced tradeoff between sensitivity and specificity. We further show that by combining clinical knowledge with a BRF, we can boost the performance and reliability of the classifier as measured by the area under the receiver operating characteristic (ROC) curve.

+Equal contribution by authors.
*Research supported in part by the Canadian Institutes of Health Research. The work of L. Kanbar was supported in part by the Natural Sciences and Engineering Research Council of Canada. K. A. Brown was supported in part by the Queen Elizabeth Hospital of Montreal Foundation Chair in Pediatric Anesthesia.

L. J. Kanbar and R. E. Kearney are with the department of Biomedical Engineering, McGill University, Montreal, QC H3A 2B4, Canada.
W. Shalish and G. M. Sant'Anna are with the department of Neonatology, McGill University, Montreal, QC H3A 2B4, Canada.
C. C. Onu and D. Precup are with the department of Computer Science, McGill University, Montreal, QC H3A 2B4, Canada.
K. A. Brown is with the department of Anesthesia, McGill University Health Center, Montreal, QC H3A 2B4, Canada.

The objective of this study was to use automated, objective features of cardiorespiratory variability to predict extubation readiness using nonlinear machine learning tools in a database of 189 extremely preterm infants. The rest of the paper is organized as follows: Section II describes the APEX study and the signals acquired; Section III details the methods used for feature development; Section IV describes the machine learning methods used to develop a predictor; Section V reports the results of the work; and Section VI provides a discussion and concluding remarks.

## II. APEX STUDY DESIGN

### A. Infant Population

The APEX study is an ongoing multicenter, prospective, observational study. Over a period of 4 years, data have been acquired from five NICUs: the Royal Victoria Hospital; the Montreal Children's Hospital; the Jewish General Hospital (Montreal, Quebec, Canada); the Detroit Medical Center (Detroit, MI, USA); and the Women and Infant's Hospital (Providence, RI, USA). Ethics approval was obtained from the Institutional Review Board of each institution.

Infants with Birth Weight (BW) ≤ 1250g, receiving Invasive Mechanical Ventilation (IMV) at the time of enrolment and undergoing their first extubation attempt were eligible. Infants were excluded if they had any major congenital anomalies, or were receiving any vasopressor or sedative drugs at the time of extubation. Written informed parental consent was obtained prior to enrolment. The attending clinician was responsible for determining extubation readiness, and data were collected immediately prior to extubation.

### B. Data Acquisition

Five cardiorespiratory signals were acquired from each infant: uncalibrated Respiratory Inductance Plethysmography (RIP) signals from bands placed around the infant's ribcage (RCG) at the level of the nipple line and abdomen (ABD) at 0.5cm above the umbilicus (Viasys ® Healthcare, USA); electrocardiography (ECG) using three electrodes (Vermed, USA, © 2010); photoplethysmography (PPG) and oxygen saturation (SAT) signals from a pulse oximeter (Masimo Radical ®).

These signals were sampled continuously during 50-60 minutes of IMV followed by 5 minutes of Endotracheal Tube Continuous Positive Airway Pressure (ETT-CPAP). Following this, the patient was extubated. Signals were anti-alias filtered at 500Hz and sampled at 1000Hz using the PowerLab 16/30 analog-digital data acquisition system (ADInstruments, Bella Vista, Australia, © 2009) with a 16-bit analog-to-digital resolution.

Also, clinical variables were collected from birth until hospital discharge and include demographic information, ventilator information, post-extubation support, and final outcomes.

## III. METHODS

This exploratory study used several methods to compute cardiorespiratory variability in order to determine the discriminatory features.

### A. Cardiorespiratory Metrics

To obtain moving measurements of cardiorespiratory behavior, the signals were processed at every time instant into sample-by-sample metrics of power, respiratory frequency, cardiac frequency, and thoraco-abdominal synchrony, as described in [11, 13]. The metrics computed for this study include:

1. Pause power in the RCG ($rp^{rc}$) and ABD ($rp^{ab}$): the power in the 0-2Hz band in a short sliding window relative to the median power in a preceding long window.

2. Respiratory frequency ($rf^{ab}$): the frequency (in a sliding window) at which the highest power occurs in the 0-2Hz band, using a bank of band-pass filters with 0.2Hz bandwidth.

3. Cardiac frequency using the ECG ($cf^{ec}$) or PPG ($cf^{pp}$): the frequency with the most power in the 1.5-3.5Hz band, using the Short Time Fourier Transform (STFT).

4. Root-mean-square ($rms^+$): the sum of the RMS of the RCG and ABD in sliding windows.

5. Thoraco-abdominal phase (Φ): the phase difference between the RCG and ABD.

6. Movement artifact power in the RCG ($bmp^{rc}$) and ABD ($bmp^{ab}$): the power in the 0-0.4Hz movement artifact band relative to the 0.4-2Hz breathing band.

7. Cross-Correlation coefficient between the cardiac frequency and respiratory frequency ($\rho_0^{rf-cf}$), computed over a sliding window.

### B. Cardiorespiratory Patterns

The RCG and ABD signals were also processed using an Automated Unsupervised Respiratory Event Analysis (AUREA) algorithm to extract the following sample-by-sample respiratory patterns:

- Pause (*PAU*): cessation of breathing indicated by low RCG and ABD power in the breathing band (0.4-2Hz).

- Movement Artifact (*MVT*): periods during which there is power in the movement artifact band (0-0.4Hz) due to infant movement or nurse handling.

- Synchronous Breathing (*SYB*): periods during which RCG and ABD are in synchrony.

- Asynchronous Breathing (*ASB*): periods during which RCG and ABD are out of synchrony.

AUREA is objective, repeatable, and has been tuned for this population. Further details are given in [13]. The following patterns were also extracted from the ECG and PPG signals:

- Bradycardia (*BDY*): artifact-free periods during which and the heart rate was below 100 beats/min.

- Desaturation (*DST*): artifact-free periods during which the oxygen saturation was less than 85%. Moving artifact was detected using a PPG movement artifact detector [14].

*C. Cardiorespiratory Features*

Cardiorespiratory variability was analyzed using the behavior provided by the metrics, the patterns, and the R-peak intervals of the ECG. Table 1 summarizes the features.

For each metric, the moving average power was computed during the 2nd minute of the ETT-CPAP period as our previous work had demonstrated that this time period was able to significantly differentiate between success and failure [11]. The 1st minute was discarded as infant's transition time to ETT-CPAP. The median and inter-quartile range (IQR) of the metrics and their power were used as scalar representations. Scalar properties of the SAT signal were also computed using median power, IQR of the power, kurtosis, and skewness.

The moving power of the heart rate was computed during the 2nd minute of ETT-CPAP. Also, R-peaks in the ECG were detected using the Pan Tompkins algorithm [15]. Several heart rate variability features were computed: the standard deviation of the time interval between R-peaks (SDNN); the standard deviation of successive differences in the interval between R-peaks (SDSD); and the triangular index [16]. The features were computed during IMV and ETT-CPAP.

Cardiorespiratory pattern variability provided a measurement of the overall performance of an infant during the ETT-CPAP period. Pattern variability for the 6 patterns was computed using the following:
1. Number of occurrences ($N^P$).
2. Total duration ($T_{tot}^P$).
3. Maximum length ($T_{max}^P$)
4. Pattern density ($D^P$): defined as the fraction of the ETT-CPAP time spent in a pattern.
5. Pattern frequency ($F^P$): defined as the number of pattern occurrences divided by the total duration of ETT-CPAP.

## IV. MACHINE LEARNING APPROACH

*A. Features*

A total of 77 cardiorespiratory variability and 2 clinical features were computed and used as input to the classification algorithms (Table 1). Three different classification strategies were employed to predict using this feature set.

*B. Random Forest*

For classification, we used Random Forest (RF) classifiers. As had been shown in previous work that linear classifiers are inadequate for this difficult clinical population, it was important to use a classifier with the ability to learn non-linear decision boundaries. Additionally, RFs have the additional benefit of embedded feature selection [17].

The random forest classifier is a bagging machine learning method [18] which works by training several decision trees in parallel. Each tree is trained on a subset of the examples and features in the dataset. This approach permits each decision tree to learn something new and different about the dataset. Such bagging methods also help to reduce variance and the chances of overfitting. During testing, each tree in the forest makes an independent prediction on the new example. The predictions from all trees are then averaged to obtain a single prediction for that example.

*C. Balanced Random Forest*

The Random Forest classifier, like many machine learning algorithms (such as logistic regression and support vector machines) encounters difficulty in making good predictions when the number of examples in the different classes is imbalanced. The skew in the dataset could be worsened in some or all of the subsets passed to the trees, potentially leading to trees that are only good at predicting the majority class (i.e. the success class). In this work, we addressed this challenge through random undersampling of the majority group. By this, we ensured that the subsets passed to the decision trees have equal number of success and failure examples. This type of random forest has been presented in literature as a Balanced Random Forest (BRF) [12].

*D. Clinical Decision & Balanced Random Forest*

Clinically, it is quite common that infants who are older and larger at birth tend to be extubated successfully, and that the difficulty in deciding when to extubate lies primarily in the younger and smaller infants. To analyze this empirically, we examined the gestational age as a function of the birth weight of our infant population (Fig. 1). Of the 80 babies who were at least 27 weeks old or weighed above 1000g, 76 (95%) were extubation successes. We applied a rule to encode this choice - all infants above 27 weeks or 1000g were automatically classified as success. A BRF was then trained on only the population of young and small infants. In doing this, we encode the choice of the clinician to extubate the low-risk population of older babies and focus the efforts of the classifier on the difficult, younger segment of the population. This classifier involves a 2-stage process – the clinical decision/stratification rule, followed by a BRF classifier (CD-BRF).

TABLE 1: CARDIORESPIRATORY VARIABILITY FEATURES PROVIDED TO ALL CLASSIFIERS

| Metrics | Features |
|---|---|
| $rp^{rc}, rp^{ab},$ $rf^{ab}, cf^{ec},$ $cf^{pp}, rms^+,$ $\phi, bmp^{rc},$ $bmp^{ab}, \rho_0^{rf-cf}$ | Median, IQR<br>Median power, IQR of power<br>*(40 features)* |
| **SAT** | Kurtosis, Skewness<br>Median power, IQR of power<br>*(4 features)* |
| **ECG**<br>*RR Intervals* | SDNN, SDSD, triangular index<br>*(3 features)* |
| **Patterns**<br>*PAU, MVT, ASB, SYB, BDY, DST* | $N^P, T_{tot}^P, T_{max}^P, D^P, F^P$<br>*(30 features)* |
| **Clinical Variables** | BW, GA<br>*(2 features)* |

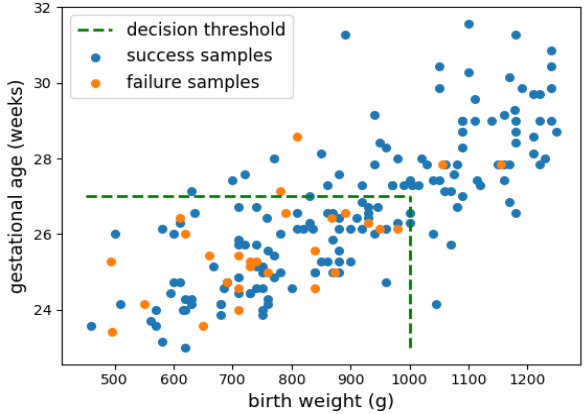

Fig. 1: Gestational Age (GA) vs Birth Weight (BW) of the patient population, showing a decision threshold separating the older, larger patients. 95% of patients above threshold were successfully extubated.

### E. Hyperparameter Search

The random forest has several hyperparameters which must be tuned appropriately for a given dataset to obtain optimal performance from the classifier. These hyperparameters include settings that affect the underlying decision trees – such as the number of features to consider at each node of the tree, the maximum depth before terminating the tree and minimum number of examples required at a leaf node – as well as settings that govern the random forest itself – such as the number of trees to train. These parameters generally control the bias-variance tradeoff between fitting an overly complex model and an overly simplistic one [17]. We employed 5-fold cross validation to evaluate our 3 classifiers (RF, BRF and CD-BRF) trained at several combinations of these hyperparameters. The best hyperparameter setting was chosen as that which gave the best performance on the test set averaged over all folds.

### F. Performance Metrics and Evaluation

For each model (i.e. hyperparameter setting) trained, we tracked 3 performance metrics: *sensitivity* or the success detection rate, *specificity* or failure detection rate and *balanced accuracy*[1], which is the average of the two. We selected the model that gave the best balanced accuracy. The receiver operating characteristic (ROC) curve and the area under the curve (AUC) were calculated for the selected model. The AUC was generated by varying a threshold on the average predicted probability from all trees.

To use our relatively small dataset as efficiently as possible, we did not leave out a fixed test set. Instead we used 5-fold cross validation to evaluate the generalization capability of our selected model. This enables us to use each example at least once for both training and testing. Performance metrics are always reported on the test set.

## V. RESULTS

At the time of this work, the database contained 189 patients with BW 882±201g and GA 26.5±1.9 weeks. Infants were extubated at 13.3±15.6 days post-birth when deemed 'ready' by the attending physician. A total of 28 (14.8%) infants failed extubation. The dataset contained 189 patients – 161 succeeded extubation and 28 failed, i.e., required reintubation within 72hrs. The features described in Table 1 were used in the analyses. For the 6 patients for whom 3 feature values were missing, the values were imputed using the median of the values from other patients in the same group.

Fig. 2 shows the performance of the 3 classifiers: random forest (RF), balanced random forest (BRF), and clinical decision with balanced random forest (CD-BRF) as ROC curves. First, it can be seen that the CD-BRF performs best, attaining the highest AUC (0.74) among all 3 classifiers. Second, whereas the standard random forest classifier (RF) learns a skewed model with a high false positive rate (as seen in the low 43% specificity), BRF and CD-BRF use random undersampling of the majority class and attain a better balance between true positive and false positive rates (with specificity of 75% and 71% respectively). It is worth noting that the CD-BRF classifier achieved an 83% specificity in the younger population denoted by Fig. 1.

The features selected by the CD-BRF classifier were examined. It was found that only 17 of the 77 cardiorespiratory variability features had non-zero weights, suggesting that the remaining 60 features had no correlation with the outcome. The top six features and their weights are shown in Table 2. Future work will evaluate how these features differ in the two groups of patients and a determination of what new, potentially useful features can be included to boost performance.

## VI. DISCUSSION

This paper presented an approach for predicting extubation readiness from automated, novel cardiorespiratory variability features using non-linear classifiers based on random forests. This work extended upon the pilot work published previously by incorporating novel features sensitive to breathing patterns in a larger, multi-institutional population. Class imbalance was accounted for by randomly undersampling the majority class in a balanced random forest (BRF). Our best classifier combined clinical domain knowledge with a BRF to give a success detection rate of 78% and failure detection rate of 71%. This suggests that there were signs in the cardiorespiratory behavior of these infants which, if considered by the physicians, could have prevented 71% extubation failures.

Previous work using cardiorespiratory variability features achieved a failure detection rate of 83.2% and success detection rate of 73.6% [11]. This was carried out on a much smaller sample size of 53 babies. The performance observed in current work may be a more realistic measure given the increased heterogeneity in the population. In this work, the best AUC of 0.75 compares with that of [19] which used only clinical variables, and is better than results in [20] which used only respiratory patterns. Overall, this highlights the difficulty of predicting extubation readiness in such a high-risk population.

---

[1] The concept of *balanced* accuracy is different from *balanced* random forest. See appendix in [20] for a motivation on why balanced accuracy is a preferred metric over accuracy.

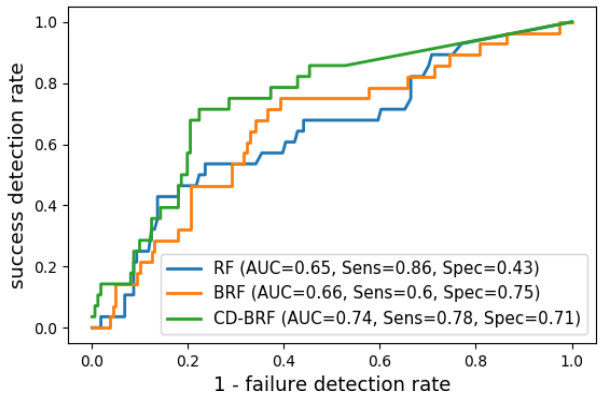

Fig. 2: ROC curve for standard random forest (RF), balanced random forest (BRF) and classifier which combines clinical decision with balanced random forest (CD-BRF). AUC – area under the curve. "Sens" and "Spec" refer to the sensitivity (success detection rate) and specificity (failure detection rate) at the best point on the ROC.

There were a few limitations with this work. First, we reduced the time-varying signals into scalar features in order to leverage them in the classifiers. These scalar representations may have discarded useful information in the signals. In future work, it will be necessary to experiment with machine learning methods that are inherently designed for time-series data, such as hidden Markov models (HMM) and conditional random fields (CRF). Second, only the 2nd minute of the ETT-CPAP was considered for consistency and to allow direct comparison with previous work. Given that only 17/77 features showed importance, it is crucial to explore new features or longer ETT-CPAP periods.

Also, clinical variables such as the weight and age at extubation may also contribute to a better stratification in the CD-BRF classifier. Future work will examine these possibilities. Finally, as the number of failure cases was quite small, it will be important to test the models developed here on a held-out validation set. Indeed, this is part of our data acquisition protocol and will be tested in future work.

ACKNOWLEDGMENT

We thank the patients and their families for their participation in this research.

TABLE 2: TOP 6 FEATURES SELECTED BY CD-BRF CLASSIFIER.

| Feature Name | CD-BRF Weight |
|---|---|
| Median of $bmp^{rc}$ | 0.11 |
| $N^{ASB}$ | 0.11 |
| IQR of $bmp^{rc}$ | 0.10 |
| Median Power of SAT | 0.08 |
| IQR of $rp^{rc}$ | 0.07 |
| $F^{ASB}$ | 0.07 |